ICDSST 2020 PROCEEDINGS – ONLINE VERSION
THE EWG-DSS 2020 INTERNATIONAL CONFERENCE ON DECISION SUPPORT SYSTEM TECHNOLOGY
I. Linden, M.T. Escobar, A. Turón, F. Dargam, U. Jayawickrama (editors)
Zaragoza, Spain, 27-29 May 2020

# Selection of an Integrated Security Area for locating a State Military Organization (SMO) based on group decision system: a multicriteria approach

Jean Gomes Turet[1,2], Ana Paula Cabral Seixas Costa[1] and Pascale Zaraté[2]

[1]Universidade Federal de Pernambuco

Management Engineering Dep., Av. da Arquitetura, S/N, 50740-550, Recife-PE, Brazil

[2] Toulouse University, Université Toulouse 1 Capitole, 2 Rue du Doyen-Gabriel-Marty 31042 Toulouse, France.

IRIT- UMR 5505.

## ABSTRACT

Over the past few years there has been growing concern among authorities over crimes committed worldwide. In Brazil it is no different. High crime rates have encouraged government authorities involved in public safety to identify solutions to minimize crimes. In this context, one way to plan and manage security is in the division of neighborhoods in ISA (Integrated Security Areas). Each ISA has a neighborhood conglomerates taking into account their geolocation. From this it becomes possible to maximize security management and combat crime. Based on that, one of the main points that generate great discussion at the governmental level is the choice of a certain integrated security area for the installation of a certain police battalion. This choice involves multiple decision makers since several hierarchies are involved. Thus, this paper aims to identify the best ISA to deploy a police battalion using group decision techniques and tools. For this work the Group Decision Support System (GDSS) called GRoUp Support (GRUS) was used from two main Vote techniques: Condorcet and Borda. With this it was possible to identify the best ISA taking into account the pre-established criteria.

**INTRODUCTION**

Public safety has received special attention in many countries, especially regarding lack of security. Particularly in Brazil, there are several problems that contribute to a lack of security. Low investment in technologies, inefficient integration between the various public safety departments, and low salaries are among the points that have the greatest impact on the effectiveness of public policies for the fight against crime [1][2]. The state of Pernambuco,

Brazil, uses an index that identifies the number of victims, the Violent Lethal and Intensive Crimes (VLIC), and this index shows worrying data. The Social Defense Department of Pernambuco (SDS PE) integrates actions of the Pernambuco government aimed at guaranteeing public order to minimizing the VLIC index, considering the population's feelings towards the actions that are carried out, as well as the sense of security of each individual.

One of the points that have the biggest effect the individual sense of security is violence against life. These crimes have particularly strong effects on the daily life of people in developing countries, as is the case of Brazil [2]. Most of these crimes are some form of theft followed by murder, which causes profound impacts in both social and cultural terms. Specifically, in social terms, it is evident that the sentiment of the population is affected by a new crime. This feeling is characterized by the fear that people have about the possibility of suffering a similar crime [2][3]. Thus, programs for the promotion of public safety must both minimize crime and increase the sense of security in the population.

In this context, in order to manage public security, the state government of Pernambuco, together with the social defense department, divides neighborhoods and cities from integrated security areas (ISA). Each ISA is responsible for a set of neighborhoods enabling greater management of public safety. From this division it becomes possible to expand the offer of actions that aim to minimize violence and, consequently, increase the sense of security.

However, one of the most critical points of public security management is the installation of new police stations, as there are several criteria that need to be analyzed, as well as multiple decision makers with different profiles, as well as taking into account political and social factors.

Thus, this paper aims to identify the best ISA to deploy as a police battalion using group decision techniques and tools. For this work we used the GRUS software from two main Vote techniques: Condorcet and Borda. With this it will be possible to identify the best ISA taking into account the pre-established criteria.

**BACKGROUND**

**Integrated Security Areas (ISA)**

In Pernambuco there are 26 integrated security areas (ISA). These are smaller areas that are used to define how the police should work, allowing for greater level of cooperation and integration between the parties involved (Figure 1). From this integration it becomes possible to establish actions aimed at minimizing the occurrence of crimes, such as homicides.

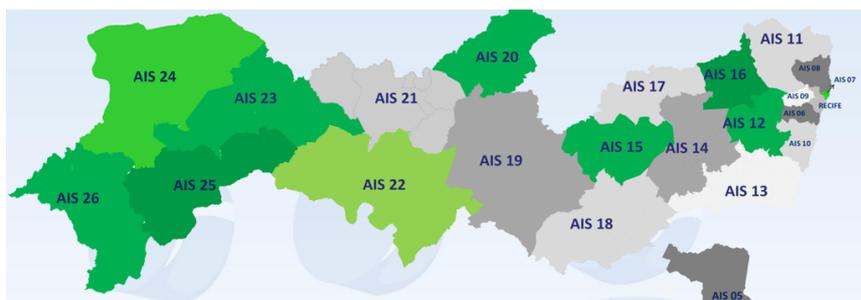

Figure 1: Integration Security Areas (ISA)

These 26 areas are distributed into three Regional Directors. The first group is subordinate to the Director of the Metropolitan Region and is composed of police stations located in the state capital and cities of the metropolitan region; The second group reports to the Director of Interior Area 1, which involves cities in the central region of the state. and the third group reports to the Interior Area Director 2, who considers cities in the western state. Although these 26 ISAs exist, the crime rate per population, which is above reasonable limits, is a major problem in some municipalities in the state of Pernambuco.

**Police Station location problems**

According to [4], localization problems are fully associated with geography, in which the division of an area by a police force is fundamentally a geographical problem. Generally, a city is divided into police command and patrol areas. [5] According to the author, the boundaries of an area are based on expert knowledge of the areas to be patrolled and police resources.

As stated earlier, optimizing the number of police stations and the allocation decision making problem should consider a different set of criteria. Based on the literature [4] [6] there are several suggested criteria that can be considered, such as the crime rate per population (one of the important criteria for police station optimization); population and unemployment rate; identification of a potential crime center; road accessibility and land availability; population density; Size of territory an agent should patrol; potential growth; density of police stations; political considerations; and environmental impact.

The State of Pernambuco is divided into Integrated Security Areas in a manner similar to that defined by [10], in a real scenario, decisions to define the location of an **State Military Organization** (SMO) are based basically on political and / or financial issues and without support for a formal procedure (subjective analysis). Thus, the contribution of this paper is to support this kind of decision problem, considering a relevant set of decision maker criteria and preferences.

**GRUS System (Condorcet and Borda)**

GRUS software is equipped with various collaborative tools and can be used in various decision-making processes by defining steps that make up this process, such as Brainstorming - Categorization of ideas - Establishment of consensus that is part of a group decision process. If there is no consensus, voting processes are required [7]

For the Borda process, the candidates are sorted according to the preferences of each voter. A count is then proceeded follows: each position in the ranking is given a score: 1 point for the last ranked, 2 for the second to last, 3 for the second to last etc., ie the distance between each preference should be only one point. In the end the points are added to decide which alternative wins. This system takes into account not only each voter's first choice, but all others, so the winner is not always the first-placed candidate [8]

Regarding the Condorcet methods, or parity methods, are a class of ordered voting methods that follow the Condorcet criteria. These methods compare every pair of options, and the option that beats every other option is the winner. One option outperforms another option if most votes rank it better on the ballots than the other option. These methods are often referred to collectively as Condorcet methods, because the Condorcet criterion ensures that they all give the same result in most elections, where there is a Condorcet winner [8].

**METHODOLOGY**

The evaluation matrix is available thanks to a previous work [9]. In this paper the authors organized the questionnaire and a workshop where three stakeholders were involved. Based on interviews and the consolidation of opinions, the financial, social, criminal and political aspects were highlighted by the experts as important for achieving better and more compromising solutions. These four aspects were also fragmented into more specific issues: intentional lethal violent crime (ILVC), heritage violent crime (PVC), hotspot analysis, territorial extension, commercial presence, number of prisons, significant presence of communities, area with presence of flow, population, number of schools, risk assessment area. From this we identified the weights based on the importance of preferences among the stakeholder criteria (Table 1).

Table 1: Criteria

| Criteria | Scales | Weights |
|---|---|---|
| CVLI (c1): | Total amount of intentional lethal violent crimes | 0.30 |
| Population (c2): | Population | 0.20 |
| Number of prison (c3): | Number of prisons in the area | 0.10 |
| Presence of another (SMO) (c4): | Number of SMO in the area | 0.10 |
| CVP rate (c5) | Total amount of patrimonial violent crime | 0.15 |
| Perceived risk level (c6) | 5-point Likert scale (from very low risk to very high risk) | 0.15 |

Table II illustrates the evaluation matrix considering 26 alternatives (representing 26 ISA) as well as the values of each alternative for all criteria.

Table 2: Evaluation Matrix

| Alternative \ Criteria | C1 | C2 | C3 | C4 | C5 | C6 |
|---|---|---|---|---|---|---|
| ISA_1 | 62 | 78.097 | 0 | 1 | 8420 | 3 |
| ISA_2 | 90 | 337.983 | 0 | 1 | 7308 | 2 |
| ISA_3 | 162 | 402.419 | 0 | 1 | 8182 | 4 |
| ISA_4 | 176 | 421.291 | 2 | 1 | 9262 | 4 |
| ISA_5 | 150 | 375.081 | 0 | 1 | 4711 | 3 |
| ISA_6 | 416 | 751.002 | 0 | 2 | 9145 | 5 |
| ISA_7 | 170 | 380.556 | 0 | 1 | 8372 | 4 |
| ISA_8 | 269 | 618.090 | 5 | 1 | 7617 | 5 |
| ISA_9 | 126 | 267.570 | 0 | 1 | 3073 | 3 |
| ISA_10 | 226 | 307.271 | 0 | 1 | 2511 | 5 |
| ISA_11 | 324 | 576.958 | 0 | 2 | 3019 | 3 |
| ISA_12 | 235 | 418.012 | 1 | 2 | 2596 | 3 |
| ISA_13 | 318 | 493.076 | 1 | 2 | 1804 | 5 |
| ISA_14 | 408 | 676.723 | 1 | 1 | 7242 | 4 |
| ISA_15 | 145 | 322.263 | 1 | 2 | 1508 | 3 |
| ISA_16 | 123 | 355.549 | 1 | 2 | 1206 | 4 |
| ISA_17 | 194 | 278.246 | 0 | 1 | 2467 | 4 |
| ISA_18 | 217 | 509.005 | 1 | 1 | 1592 | 3 |
| ISA_19 | 122 | 348.158 | 2 | 1 | 500 | 2 |
| ISA_20 | 35 | 186.421 | 0 | 1 | 170 | 1 |
| ISA_21 | 71 | 186.091 | 0 | 1 | 262 | 2 |
| ISA_22 | 51 | 171.471 | 0 | 2 | 228 | 1 |
| ISA_23 | 42 | 146.341 | 1 | 1 | 258 | 2 |
| ISA_24 | 121 | 329.483 | 0 | 2 | 620 | 2 |

From these values it was possible to perform a simulation process in the GRUS software to identify which ISA will be chosen for the implementation of a new battalion. As previously stated, the software establishes a voting system where the parties involved can express their preferences taking into consideration the established criteria.

Thus, we have as final result (the first five placed) (Table 3):

Table 3: Final Result

| Integrated Security Area | Final Classification |
|---|---|
| AIS 02 | 1° |
| AIS 05 | 2° |
| AIS 24 | 3° |
| AIS 18 | 4° |
| AIS 04 | 5° |

Thus, it was found that AIS 02 was the winner taking into consideration the determined criteria, as well as the preference of the decision makers (which in this case were three).

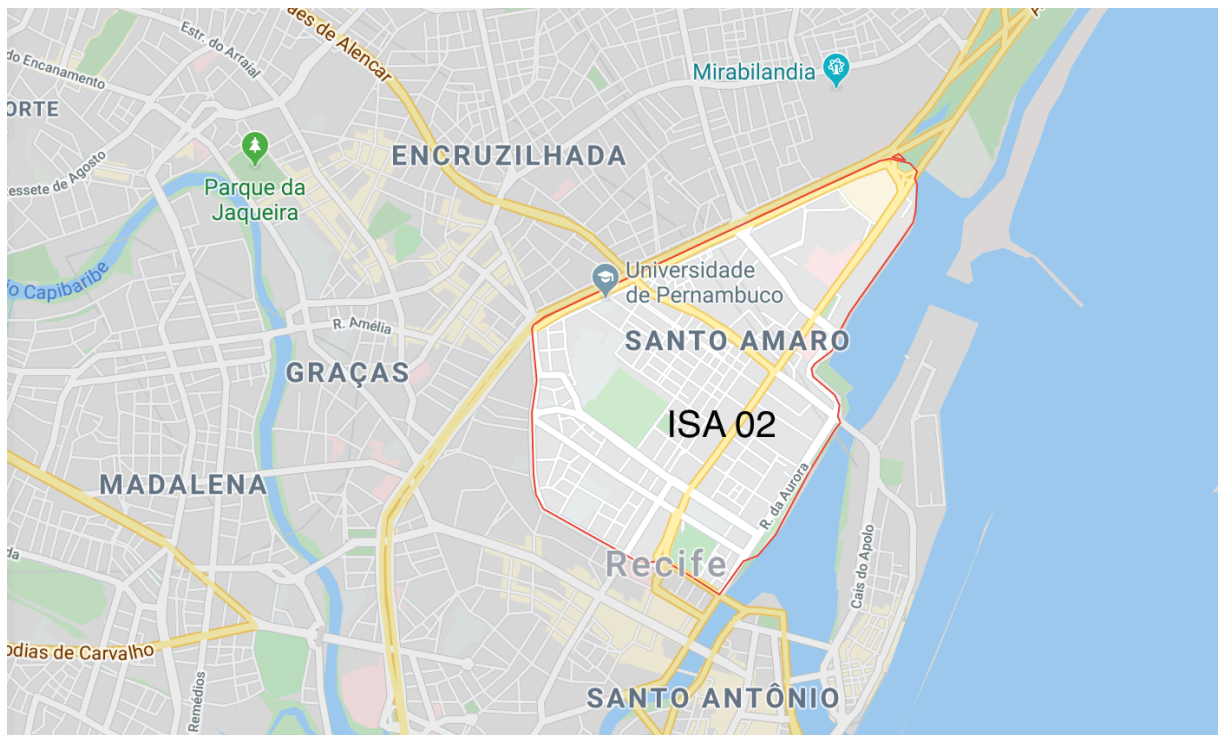

Figure 2: Integration Security Area 02

This ISA belongs to the city of Recife and is responsible for the conglomerate of three neighborhoods: Santo Amaro, Boa Vista and Soledade. In Recife, these neighborhoods represent a high crime rate considering the CVLI and CVP data of the state of Pernambuco.

**CONCLUSIONS**

Ensuring the public safety of the population is not an easy task and constantly needs actions to ensure greater efficiency in this process. In this scenario, government agencies, daily, need to make decisions that will impact the life of the population.

In this context the choice of a certain integrated security area becomes a determining factor to ensure a higher level of security in the locality and, consequently, in the state. However, this becomes a challenge as there are several criteria and decision makers who have different views of the same goal.

Thus, this work, based on the GRUS software, established a voting process taking into consideration the Condorcet vote procedure for the selection of a certain integrated security area where a new police battalion will be deployed. Thus, the purpose of this paper is to help governmental agencies decide to install a new battalion taking into account various criteria and decision makers.

For future works, besides dealing with geographical aspects (the second step of the problem location), probably using GIS, we suggest considering other set of criteria.